\documentclass[letterpaper, 10 pt, conference]{ieeeconf}  

\IEEEoverridecommandlockouts                              

\usepackage{times}
\usepackage{multicol}
\usepackage[bookmarks=true]{hyperref}
\usepackage{amsmath} 
\usepackage{amssymb}  
\usepackage{multirow}
\usepackage{subcaption}
\usepackage{graphicx}
\usepackage{booktabs}
\usepackage{flushend}
\usepackage{makecell}
\usepackage{mathtools}
\usepackage{placeins}
\usepackage{cite}
\usepackage{gensymb}
\usepackage{tikz}

\title{\LARGE \bf
Online Calibration of a Single-Track Ground Vehicle Dynamics Model by Tight Fusion with Visual-Inertial Odometry 
}
\author{Haolong Li$^{1}$ and Joerg Stueckler$^{1}$
	\thanks{* This work was supported by Max Planck Society and the Cyber Valley Research Fund (project no. CyVy-RF-2019-05). 
		The authors thank the International Max Planck Research School for Intelligent Systems (IMPRS-IS) for supporting Haolong Li.
		We thank Felix Grueninger (MPI-IS) for building the robot used in our experiments.
	}
	\thanks{$^{1}$All authors are with the Embodied Vision Group,
		Max Planck Institute for Intelligent Systems, T\"ubingen, Germany
		{\tt\footnotesize \{haolong.li,joerg.stueckler\}@tue.mpg.de}}
}

\newcommand\copyrighttext{%
   \textcopyright 2024 IEEE. Personal use of this material is permitted. Permission from IEEE must be
obtained for all other uses, in any current or future media, including
reprinting/republishing this material for advertising or promotional purposes, creating new
collective works, for resale or redistribution to servers or lists, or reuse of any copyrighted
component of this work in other works. }
\newcommand\copyrightnotice{%
\begin{tikzpicture}[remember picture,overlay]
\node[anchor=south, font=\fontsize{6}{10}\selectfont, yshift=5pt] at (current page.south) {\fbox{\parbox{\dimexpr\textwidth-\fboxsep-\fboxrule\relax}{\copyrighttext}}};
\end{tikzpicture}%
}

\begin{document}

\maketitle
\begin{tikzpicture}[remember picture, overlay]
    \node [align=center, font=\fontsize{8}{10}\selectfont, yshift=-0.5cm] at (current page.north) {
    Accepted for publication in {IEEE} International Conference on Robotics and Automation (ICRA), 2024
    };
\end{tikzpicture}
\thispagestyle{empty}
\pagestyle{empty}
\copyrightnotice

\begin{abstract}

Wheeled mobile robots need the ability to estimate their motion and the effect of their control actions for navigation planning. In this paper, we present ST-VIO, a novel approach which tightly fuses a single-track dynamics model for wheeled ground vehicles with visual-inertial odometry (VIO). Our method calibrates and adapts the dynamics model online to improve the accuracy of forward prediction conditioned on future control inputs. The single-track dynamics model approximates wheeled vehicle motion under specific control inputs on flat ground using ordinary differential equations. We use a singularity-free and differentiable variant of the single-track model to enable seamless integration as dynamics factor into VIO and to optimize the model parameters online together with the VIO state variables. We validate our method with real-world data in both indoor and outdoor environments with different terrain types and wheels. In experiments, we demonstrate that ST-VIO can not only adapt to wheel or ground changes and improve the accuracy of prediction under new control inputs, but can even improve tracking accuracy.
\end{abstract}

\section{Introduction}
 Autonomous mobile robot navigation requires the ability to perceive the extrinsic environment for localization and knowledge of an accurate robot dynamics model for path planning and control. 
 Most previous works for ground robots solve the problems of state-estimation and calibration of the dynamics model of the robot drive separately. For the perception and localization part, visual-inertial odometry (VIO) methods (e.g.~\cite{leutenegger2015_okvis,qin2017vins, usenko19nfr})
 have become popular in the computer vision community due to the low-cost and outstanding tracking accuracy. On the other hand, many works from the robotics community try to estimate vehicle slip angle and vehicle parameters such as mass or tire coefficients where vehicle pose, velocity, and acceleration can be measured by GPS or other odometry methods (e.g.~\cite{reina2017_ekfparameterid, wielitzka2015_onlineparameterid, you2017_jukfparamid}). In this work, we also focus on wheeled robots and propose a VIO method that can estimate robot pose states and calibrate the dynamics model of the drive jointly. 
 
 \begin{figure}[tp!]
	\centering
	\includegraphics[width=0.89\linewidth]{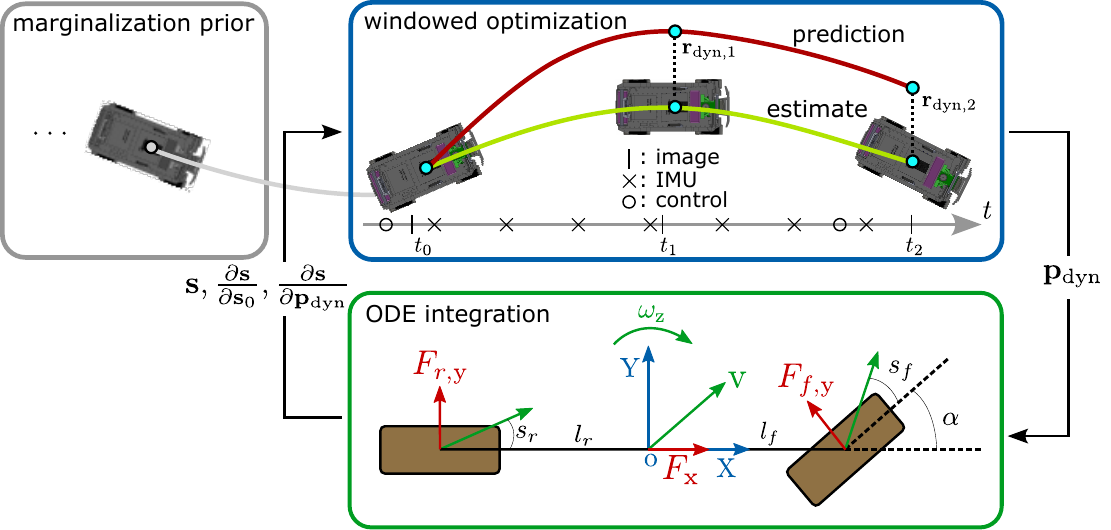}	
	\caption{
ST-VIO performs windowed optimization (blue box) with marginalization of old states (gray box) to estimate vehicle motion and parameters of a single-track dynamics model.
The dynamics model is used as factor in the optimization through ODE integration (green box, 
wheels: brown rectangles, velocity: green, force: red, x-axis: longitudinal, y-axis: lateral axis).
	}
	\label{fig:teaser}
\end{figure}

The integration of motion models into VIO has already been extensively researched. Some studies aim to use motion models with sensors such as wheel odometers to constrain the VIO and improve the tracking accuracy (e.g.~\cite{wu_vinsonwheels,lee_onlincalib}).    
In contrast, studies like~\cite{nisar_vimo,cioffi2023hdvio} integrate the dynamics models of a multicopter not just for improving tracking accuracy but also for external force prediction.
In this study, we focus on ground-based wheeled robots and incorporate a dynamics model with friction and tire-ground interactions into VIO. 
This model serves not only as a motion constraint for tracking but is also calibrated online, improving the accuracy of forward prediction based on control inputs. 
We employ the single-track model, also referred to as the bicycle model, where the left and right wheels are lumped as one (green box in Fig.~\ref{fig:teaser}). 
It provides a good trade-off between accuracy and computational efficiency~\cite{kabzan2019_racingdynamicslearning}.
The integration of this dynamics model into VIO presents significant challenges. Firstly, the model's behavior can vary due to changes in terrain properties or tire conditions. Secondly, this model encounters a singularity when vehicle speed nears zero. 
To overcome these hurdles, we modify the dynamics model to eliminate the singularity and implement real-time online parameter calibration together with VIO state variable optimization. 
Our method continuously adapts the model, enabling more accurate predictions of vehicle pose and velocity based on the latest state estimates and new control inputs.
This adaptive prediction could enable potential applications in downstream tasks such as model-predictive control and navigation planning.

We evaluate our method in robot experiments in indoor and outdoor environments. 
 Our experiments demonstrate that integrating the robot drive dynamics model can improve the tracking accuracy. 
 Moreover, the online calibration is capable of adapting the parameters so that the accuracy of prediction with the model is improved. 
 In summary, the main contributions of our work are: (1) We tightly integrate a singularity-free single-track vehicle model which is formulated as an ordinary differential equation (ODE) as a multistep motion constraint for ground wheeled-robots into VIO. This enables online real-time estimation and calibration of model parameters alongside VIO state variables. (2) We demonstrate that our method not only enhances VIO tracking accuracy but also allows the model to adapt to variations in terrain and vehicle properties.

\section{Related Work}
Several prior works exist in the literature which model wheeled vehicle dynamics and identify model parameters by matching state estimates from the model with ground truth recordings from real vehicles.
Wielitzka et al.~\cite{wielitzka2015_onlineparameterid} filter parameters of a double-track dynamics model with vehicle states including side-slip angle using an Unscented Kalman Filter based on a GPS-gyro measurement system.
A similar approach is taken in~\cite{you2017_jukfparamid} which estimates parameters of both a single-track model and an extended double-track model that models air resistance and sprung and unsprung mass separately. 
Aghli et al.~\cite{aghli_2018} identify the parameters of the robot dynamics model online using ground truth from a motion capture system. 
In our approach, we calibrate a single-track dynamics model online jointly with visual-inertial odometry state estimation.
%
Xu et al.~\cite{xu2019_largescaledynlearning} developed a learning-based approach that trains multilayer perceptrons or LSTMs~\cite{hochreiter1997_lstm} to model the dynamics.
Kabzan et al.~\cite{kabzan2019_racingdynamicslearning} and Jiang et al.~\cite{jiang2021_vehicledynamicsresidual} train a Gaussian Process or neural residual models to improve predictions of a base dynamics model.
Such data-driven methods, however, require that training and test distribution are sufficiently similar to generalize well to cases unseen during training, while our method adapts online.
Using visual-inertial sensors to estimate robot states is appealing due to their lower cost and greater flexibility compared to satellite measurements or motion capture systems.
Since the VIO system for ground robot motion is ill-posed~\cite{wu_vinsonwheels}, numerous previous studies have sought to integrate and calibrate either kinematic or dynamic motion models with VIO to enhance tracking accuracy (e.g.,~\cite{ma2019_ackmsckf,zhang2020_vinsplvehicle,xiong2021_gvido}).
These methods, however, typically do not pursue to calibrate the parameters of the motion model online for motion prediction as in our approach. 
Weydert~\cite{weydert2012_ekfvoparameterid} estimate the vehicle ego motion and model parameters with a dual ensemble EKF using stereo cameras. 
The method does not learn a forward model mapping between control commands and vehicle state like our method. 
In our previous work~\cite{li2022_viokinmodel}, we calibrate parameters of a velocity-control based  kinematic motion model online by tight integration with stereo visual-inertial odometry~\cite{usenko19nfr}. However, the kinematic motion model does not take tire-ground interaction and dynamics into consideration.
Notably, above mentioned methods such as~\cite{weydert2012_ekfvoparameterid, xiong2021_gvido} do not address low-speed scenarios due to singularities of the dynamics model at zero speed, potentially compromising consistent model calibration at low speeds. 
Zhang et al.~\cite{zhang2018_tiremodeling} propose a non-smooth model which caps velocities at low speeds.
An alternative approach mentioned in~\cite{zhang2018_tiremodeling} switches to a kinematic model which would add complexity for integration as dynamics factor and prediction. 
We adjust the dynamics model to eliminate singularities and ensure it remains differentiable. 


\section{Method}

In our approach we tightly fuse a single-track vehicle dynamics model with VIO to improve state estimation and  facilitate online calibration.
Throughout this paper, bold capital letters~(e.g., $\mathbf{R}$) represent matrices, bold lowercase letters represent vectors~(e.g., $\mathbf{v}$) and non-bold letters stand for scalars~(e.g., $\gamma$). We interchangeably denote the rigid body pose as $\mathbf{T} \in \mathrm{SE(3)}$ or $(\mathbf{R} \in \mathrm{SO(3)}, \mathbf{p}\in \mathbb{R}^{3})$.
The origin of the world frame $\mathrm{w}$ is set to the initial position of the camera, and its z-axis is upwards aligned with gravity.

\subsection{Single-Track Dynamics Model}
\label{sec:singletrack}

The single-track vehicle dynamics model is commonly used for navigation of ground wheeled robot due to its balance of simplicity and accuracy~\cite{kabzan2019_racingdynamicslearning}. 
As depicted in the green box of Fig.~\ref{fig:teaser}, the local body frame $\mathrm{o}$ locates at the center of mass of the vehicle and is assumed to be fixed. The x-axis of the body frame points forward and the z axis~(yaw axis) points upward.
We define the state variables of the dynamics system in the body frame $\mathrm{o}$ as
	$\mathbf{s} = \begin{pmatrix}
	x, & y, & \theta, & v_\mathrm{x}, & v_\mathrm{y}, & \omega_\mathrm{z}
	\end{pmatrix}  ^\top$, 
where $x$ and $y$ are the 2D position, and $\theta$ is the yaw rotation along the z-axis. The velocities $v_\mathrm{x},v_\mathrm{y}$ are the corresponding linear velocities and $\omega_\mathrm{z}$ is the yaw velocity.
The control inputs of the dynamics system are the throttle control $u_{\mathrm{thr}} \in [0, 1] $ and steering control $u_\mathrm{str} \in [-1, 1]$.
The dynamics system itself is an ordinary differential equation~(ODE) system expressed as
\begin{multline}
    \dot{\mathbf{s}} = \left(
    v_\mathrm{x} - \omega_\mathrm{z} y,
    v_\mathrm{y} + \omega_\mathrm{z} x,  
    \omega_\mathrm{z}, 
    \frac{F_{\mathrm{x}} - F_{f,\mathrm{y}} \sin(\alpha)}{m} + v_\mathrm{y} \omega_\mathrm{z},\right.\\
    \left.\frac{F_{f,\mathrm{y}} \cos(\alpha) + F_{r,\mathrm{y}}}{m} - v_\mathrm{x} \omega_\mathrm{z}, 
    \frac{l_f F_{f,\mathrm{y}} \cos(\alpha) - l_r F_{r,\mathrm{y}}}{I_\mathrm{z}}
    \right)^\top  \hspace{-0.7ex}
\end{multline}
where $m$ and $I_\mathrm{z}$ are the vehicle mass and yaw momentum of inertia.
As illustrated in the green box of Fig.~\ref{fig:teaser}, $l_{f/r}$ represent the distances from the front and rear wheels to the center of mass, respectively, $\alpha$ denotes the front wheel angle,~$F_{\mathrm{x}}$ is the longitudinal force at the center of mass in the body frame depending on the throttle input, while $F_{f/r,\mathrm{y}}$ refer to the lateral tire forces at the front and rear wheels.

We write the parameters that we aim to calibrate online as a vector
	$\mathbf{p}_{\mathrm{dyn}} = \begin{pmatrix}\gamma, & C_{\mathrm{thr},1} & 
	 C_{\mathrm{thr},2} &  C_{\mathrm{res}}, & C_{\mathrm{tire}} \end{pmatrix} ^\top$.	
The first term $\gamma$ represents the steering ratio between the front wheel angle and the steering input as $\alpha = \gamma u_{\mathrm{str}}$. $C_{\mathrm{thr},1/2}$ and $C_{\mathrm{res}}$ are longitudinal force related parameters: $F_\mathrm{x} = f(C_{\mathrm{thr},1} u_{\mathrm{thr}} - C_{\mathrm{thr},2} v_\mathrm{x})  - \tanh(\sigma v_\mathrm{x}) C_{\mathrm{res}} $. The first term approximates the power-train force, which is a non-linear function of throttle input, motor speed, and other effects~\cite{LYNCH2016399}.  
We empirically approximates the non-linearity for our motor with function $f(x) = \psi x + \tau \log(1 + \exp(x)) - \log(2)$, that scales up the acceleration force and scales down the deceleration force of the motor. The hyper-parameters~$\psi$ and~$\tau$ control the scaling ratio.  
We ignore the air drag force and model the resistance as a scalar $C_{\mathrm{res}}$, which is multiplied
with a hyperbolic tangent function $\tanh(\sigma v_\mathrm{x})$ such that no false longitudinal force will be applied when the vehicle stands still and no throttle input is given. Here, $\sigma$ is a hyper-parameter that controls the steepness of the hyperbolic tangent around zero. The hyper-parameters related to the longitudinal force, $\{ \psi, \tau , \sigma \}$, are optimized offline as explained in a later section.
Lastly, $C_{\mathrm{tire}}$ is the tire coefficient of a linear tire model for the lateral tire force $F_{f/r,\mathrm{y}}$ as explained next.

Similar to~\cite{kabzan2019_racingdynamicslearning}, 
we estimate the lateral tire force from the front and rear slip angle $s_{f/r}$, as depicted in Fig.~\ref{fig:teaser}. They are the difference between the wheel angle and wheel velocity angle and can be computed as  
\begin{align}
  s_f &= \arctan{\frac{v_\mathrm{x} \sin (\alpha) - (v_\mathrm{y} + l_f \omega_\mathrm{z}) \cos (\alpha)}            {g( {v_\mathrm{x} \cos (\alpha) + (v_\mathrm{y} + l_f \omega_\mathrm{z}) \sin (\alpha)} )  }}, \\
  s_r &= \arctan{\frac{l_r \omega_\mathrm{z} - v_\mathrm{y}} {g(v_\mathrm{x})} },
\end{align}
where $g(x) = x$ in the original model.
They are undefined at zero longitudinal velocity, $v_\mathrm{x}=0$. In~\cite{zhang2018_tiremodeling}, this singularity is handled by setting a constant lower bound for $v_\mathrm{x}$ using a threshold function. However, this non-smoothness complicates its application in the factor graph optimization. Instead, we use a soft thresholding function $ g(x) = \log(\exp({2x}) + 1) -x$ to maintain differentiability of the model. 
We use a linear tire model for computational efficiency and compute the lateral tire forces by $F_{f,\mathrm{y}} = C_{\mathrm{tire}} s_f, F_{r,\mathrm{y}} = C_{\mathrm{tire}} s_r$.



\subsection{Integration of Single-Track Dynamics with VIO}
\begin{figure}[tb]
	\centering
	\includegraphics[width=0.89\linewidth]{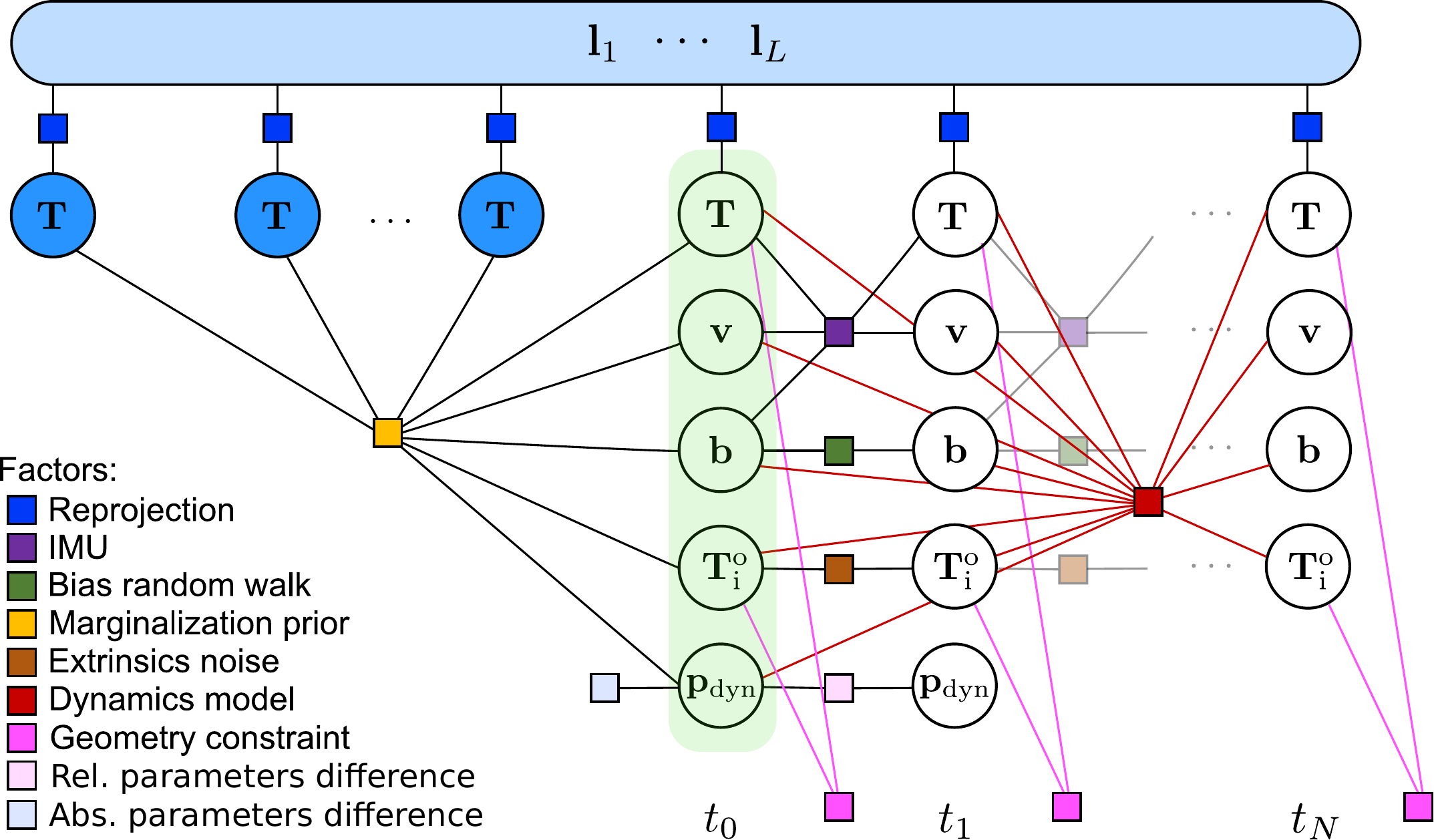}	
	\caption{
	Factor graph of ST-VIO. Blue/white circles: keyframe/recent frame variables; light green: first active recent frame at~$t_0$.
	The dynamics factor (red) connects poses, velocities, gyroscope biases, extrinsic poses of all active recent frames
	and the dynamics model parameters at $t_0$. 
	}
	\label{fig:graph}
\end{figure}
The factor graph of the windowed optimization of our ST-VIO is depicted in Fig.~\ref{fig:graph}. Within each window, there's a collection of landmarks $\mathbf{l}$, keyframes, and active recent frames. In the base VIO~\cite{usenko19nfr}, the keyframe's state is the pose of the IMU in the world frame $^{\mathrm{w}}\mathbf{T}_{\mathrm{i}, t}$, the recent frames' states consist of frame pose, linear velocity in world frame and IMU biases. The VIO method optimizes these state variables by minimizing the reprojection residual between landmarks and detected keypoints in the image frames, the relative pose residual between consecutive recent frames using IMU measurements, as well as the changes of the IMU biases assuming random walk noise. As the window shifts to the subsequent timestamp, all state variables from the oldest recent frame are marginalized unless chosen as a keyframe. The marginalized data is retained as the marginalization prior.  
Due to space limitations, we kindly refer the readers to~\cite{usenko19nfr} for more details.
In our ST-VIO, we expand the recent frames' state  by incorporating the extrinsic pose from the vehicle body frame to the IMU frame, as well as the dynamics model parameters.  As a result, we model both elements as time-varying to accommodate changes in suspension and environmental conditions, respectively.
In previous works (e.g.,~\cite{ma2019_ackmsckf,zhang2020_vinsplvehicle,xiong2021_gvido}), the motion constraints are computed between every two consecutive frames using the state estimates and the parameters stored in the first one of them. Since we are interested in multistep predictions into the future, we compute the multistep motion constraint with the state estimates and the parameters $\mathbf{p}_{\mathrm{dyn}, t_0}$ stored at the first active recent frame~(denoted as at $t_0$) in the current optimization window.  As shown in Fig.~\ref{fig:graph},  the dynamics factor thus connects all recent frames in the current window.


For a time interval between two frames $[t_n , t_{n+1}]$, we solve the ODE via the Runge–Kutta method based on the most recent control input, and set the initial state as $\mathbf{s}(t_n) = \left( 0, 0, 0, v_{\mathrm{x}, t_n}, v_{\mathrm{y}, t_n}, \omega_{\mathrm{z}, t_n} \right)^\top$.
If $n = 0$, the initial velocity in body frame is computed from the VIO linear velocity estimate $^\mathrm{w}\mathbf{v}$, gyroscope measurement $^{\mathrm{i}}{\boldsymbol{\omega}}$ and rotation $^{\mathrm{o}}\mathbf{R}_{\mathrm{i}} \in \mathrm{SO}(3)$ of the extrinsics transformation~$^{\mathrm{o}}\mathbf{T}_{\mathrm{i}}$: 
$^{\mathrm{o}}\boldsymbol{\omega}_{\mathrm{z}} = \left({^{\mathrm{o}}\mathbf{R}_{\mathrm{i}}} ({{^{\mathrm{i}}\boldsymbol{\omega}} - \mathbf{b}_{g})}\right)_{\mathrm{z}}$, 
$^{\mathrm{o}}\mathbf{v}_{\mathrm{x,y}} = \left({^{\mathrm{o}}\mathbf{R}_{\mathrm{w}}} {^{\mathrm{w}} \mathbf{v}} + {^{\mathrm{o}}{\mathbf{t}_{\mathrm{i}}}} \times {^{\mathrm{o}}\boldsymbol{\omega}}\right)_{\mathrm{x,y}}$.
If the frame at $t_n$ is not the first recent frame in the current window, $v_{\mathrm{x}, t_n}, v_{\mathrm{y}, t_n}, \omega_{\mathrm{z}, t_n}$ are set to the velocity solution of the previous time interval.
The control input can also appear in between $[t_n , t_{n+1}]$. In this special case, we first solve the ODE based on the control before $t_n$ until this new control input. 
From the intermediate solution we solve the ODE again based on the new control input until $t_{n+1}$.

The numerical solution of the ODE yields
$\mathbf{s}_{t_{n+1}}$
which contains the relative 2D pose $\left(
x_{t_{n+1}},   y_{t_{n+1}},   \theta_{t_{n+1}}
\right) ^\top$
between $t_n$ and $t_{n+1}$, and the 2D velocity $\left( v_{\mathrm{x},{t_{n+1}}},   v_{\mathrm{y},{t_{n+1}}},   \omega_{\mathrm{z},{t_{n+1}}}
\right )  ^\top$
in the local vehicle body frame $\mathrm{o}$ at the time $t_{n+1}$. The relative pose between $t_0$ and $t_{n+1}$, namely the multistep prediction of the dynamics model, can be computed as
{
\setlength{\arraycolsep}{1pt}
\begin{equation}
	\left(\!\begin{matrix}
	^{t_0}x_{t_{n+1}}\\
	^{t_0}y_{t_{n+1}}\\
	^{t_0}\theta_{t_{n+1}}
	\end{matrix}\!\right) =
	\left(\!\begin{matrix}
	\cos{(^{t_0}\theta_{t_{n}})} x_{n+1} - \sin{(^{t_0}\theta_{t_{n}})} y_{n+1} + {^{t_0}x_{t_{n}}} \\
	\sin{(^{t_0}\theta_{t_{n}})} x_{n+1} + \cos{(^{t_0}\theta_{t_{n}})} y_{n+1} + {^{t_0}y_{t_{n}}} \\
	^{t_0}\theta_{t_{n}} + \theta_{n+1}
	\end{matrix}\!\right)
	\label{eq:pose_pred}
\end{equation}
}
To compare the 6-DoF camera motion estimate of the VIO with the 3-DoF ground motion prediction by the motion model, we need to transform and project the VIO estimate into the vehicle body frame. 
The first step is to compute the 6-DoF relative pose between two timestamps e.g. $t_0$ and $t_{n+1}$ in body frame~$\mathrm{o}$ from VIO estimates and extrinsics:
$
   ^{\mathrm{o},t_0}\mathbf{T}_{\mathrm{o},t_{n+1}} =  {^{\mathrm{o},t_0}\mathbf{T}_{\mathrm{i},t_0}} \left({^\mathrm{w}\mathbf{T}_{\mathrm{i},t_0}}\right)^{-1} {^\mathrm{w}\mathbf{T}_{\mathrm{i},t_{n+1}}} \left({^{\mathrm{o},t_{n+1}}\mathbf{T}_{\mathrm{i},t_{n+1}}}\right)^{-1}.
$
Then we map the 6-DoF relative pose to 3-DoF by taking only the x and y component of the translation~$ ^{\mathrm{o},t_0}\mathbf{p}_{\mathrm{o},t_{n+1}}$ and z component of the rotation~$^{\mathrm{o},t_0}\mathbf{R}_{\mathrm{o},t_{n+1}}$
\begin{equation}
    \begin{pmatrix}
        ^{t_0}\tilde{x}_{t_{n+1}}\\
        ^{t_0}\tilde{y}_{t_{n+1}}\\
        ^{t_0}\tilde{\theta}_{t_{n+1}}
    \end{pmatrix} = 
    \begin{pmatrix}
        \left(^{\mathrm{o},t_0}\mathbf{p}_{\mathrm{o},t_{n+1}}\right)_{\mathrm{x}} \\
        \left(^{\mathrm{o},t_0}\mathbf{p}_{\mathrm{o},t_{n+1}}\right)_{\mathrm{y}} \\
        \log\left(^{\mathrm{o},t_0}\mathbf{R}_{\mathrm{o},t_{n+1}}\right)_\mathrm{z}
    \end{pmatrix}.
    \label{eq:vio_2d}
\end{equation}
We penalize the difference between the predicted 2D pose by the dynamics model~(Eq.~\eqref{eq:pose_pred}) and the estimated 2D pose by the VIO~(Eq.~\eqref{eq:vio_2d}) in our dynamics residual. Additionally, we compare the predicted local body velocity $v_\mathrm{x}, v_\mathrm{y}, \omega_\mathrm{z}$ and the velocity derived from VIO estimates $\tilde{v}_\mathrm{x}, \tilde{v}_\mathrm{y}$ and gyroscope measurement $\tilde{\omega}_\mathrm{z}$.
%
The dynamics objective function is
    $E_{\mathrm{dyn}} = \sum_{n \in \mathcal{N'}} \mathbf{r}_{\mathrm{dyn}, n}^\top \mathbf{\Sigma}_{\mathrm{dyn}, n}^{-1} \mathbf{r}_{\mathrm{dyn}, n}$,
where $\mathcal{N'}$ is the set of the recent frames except for the last one in the window, $\mathbf{\Sigma}_{\mathrm{dyn}, n}^{-1}$ is a diagonal weight matrix, and $\mathbf{r}_{\mathrm{dyn}, n}$ is the residual vector stacked from position, orientation, and velocity differences.

\subsection{Geometry Constraints}
Similar as in \cite{wu_vinsonwheels} we add a stochastic plane constraint into the VIO system as the single-track model depicts only planar motion. 
We assume that the trajectory of the vehicle body frame always lies on a plane in the world frame and its z-axis is always perpendicular to this plane. 
The plane residual is 
$
    \mathbf{r}_{\mathrm{plane}} = \left( (
    ^{\mathrm{w}}\mathbf{R}_{\mathrm{i}} {^{\mathrm{o}}\mathbf{R}_{\mathrm{i}}^\top} \mathbf{e}_3 )_{\mathrm{x}, \mathrm{y}}^\top, d + \mathbf{e}_3^\top ({^{\mathrm{w}}}\mathbf{p}_{\mathrm{i}} - {^{\mathrm{w}}}\mathbf{R}_{\mathrm{i}} {^{\mathrm{o}}\mathbf{R}_{\mathrm{i}}^\top} {^{\mathrm{o}}}\mathbf{p}_{\mathrm{i}} )
    \right)^\top
$
where $\mathbf{e}_3 = \left(0, 0, 1\right)^{\top}$, and
$d$ is the distance between world origin and initial vehicle body position, and ${^{\mathrm{o}}}\mathbf{p}_{\mathrm{i}}$ is the translation of the extrinsics transformation~$^{\mathrm{o}}\mathbf{T}_{\mathrm{i}}$.
Moreover, we also incorporate prior knowledge of the vehicle's geometry information.
Fig.~\ref{fig:vehicle} shows the mobile robot we use in this work. We assume that the vehicle body frame locates close to the longitudinal axis of the vehicle because the vehicle is roughly symmetric along this axis. Since the suspension does not affect the yaw rotation of the camera wrt. the vehicle body frame, we also penalize the yaw component of $^{\mathrm{o}}\mathbf{R}_{\mathrm{i}}$ with $(^{\mathrm{o}}\mathbf{R}_{\mathrm{i}} \mathbf{e}_3)_{\mathrm{y}}$, where $\mathbf{e}_3$ is the forward axis of the IMU frame. 
We assume that the lateral distance between body and IMU frame $(^{\mathrm{o}}\mathbf{p}_{\mathrm{i}})_{\mathrm{y}}$ is close to the lateral distance between the vehicle center and the IMU frame $l_\mathrm{cam, 1}$.
The longitudinal distance between body and IMU frame $(^{\mathrm{o}}\mathbf{p}_{\mathrm{i}})_{\mathrm{x}}$ is close to the sum of $l_f$ and the distance between camera and front wheel $l_\mathrm{cam, 2}$.
The geometric residual is 
$
    \mathbf{r}_{\mathrm{geom}, n} = \big(
    \mathbf{r}_{\mathrm{plane}}^\top,
    (^{\mathrm{o}}\mathbf{R}_{\mathrm{i}} \mathbf{e}_3)_{\mathrm{y}},
    (^{\mathrm{o}}\mathbf{p}_{\mathrm{i}})_{\mathrm{y}} - l_\mathrm{cam, 1},
    (^{\mathrm{o}}\mathbf{p}_{\mathrm{i}})_{\mathrm{x}} - l_f - l_\mathrm{cam, 2}
    \big)^\top,
$
where $l_\mathrm{cam, 1}$ and $l_\mathrm{cam, 2}$ can be measured in the CAD model.
The corresponding objective function is 
    $E_{\mathrm{geom}} = \sum_{n \in \mathcal{N}} \mathbf{r}_{\mathrm{geom}, n}^\top \mathbf{\Sigma}_{\mathrm{geom}, n}^{-1} \mathbf{r}_{\mathrm{geom}, n}$,
where $\mathcal{N}$ is the set of all active recent frames and $\mathbf{\Sigma}_{\mathrm{geom}, n}^{-1}$ is a diagonal weight matrix. 

\subsection{Optimization}

The above introduced dynamics factor and geometry constraints are integrated into the VIO system. As illustrated in Fig.~\ref{fig:graph}, the dynamics factor connects all recent frames in the current window, and the geometry constraint factor is added to each recent frame. 
To guarantee a smooth change of the extrinsics over time, we minimize the term $E_{\mathrm{extr}} = \sum_{n \in \mathcal{N}'} \mathbf{r}_{\mathrm{extr}, n}^\top \mathbf{\Sigma}_{\mathrm{extr}, n}^{-1} \mathbf{r}_{\mathrm{extr}, n}$, where $\mathbf{r}_{\mathrm{extr}}$ 
is the translation and rotation difference between two adjacent extrinsic pose estimates and $\mathbf{\Sigma}_{\mathrm{extr}, n}^{-1}$ is the diagonal weight matrix. 
The dynamics model parameters $\mathbf{p}_{\mathrm{dyn}, t_0}$ stored in the first recent frame at $t_0$ are used to perform multistep prediction until the end of the current window. 
We additionally include the model parameters $\mathbf{p}_{\mathrm{dyn}, t_1}$ at the second recent frame at $t_1$ in the current window into the factor graph and minimize the difference $\mathbf{r}_{\mathbf{p}_{\mathrm{dyn, rel}}}$ between $\mathbf{p}_{\mathrm{dyn}, t_0}$ and $\mathbf{p}_{\mathrm{dyn}, t_1}$.
When the first recent frame of the old window is marginalized out, 
the marginalization prior information can thus be propagated to the parameters at the first recent frame in the new window and prevent rapid change of the model parameters. Besides this relative difference term, a weak absolute prior is added for $\mathbf{p}_{\mathrm{dyn}, t_0}$ by minimizing its difference $\mathbf{r}_{\mathbf{p}_{\mathrm{dyn, abs}}}$
to the most recent marginalized parameters to avoid drift when the parameters become unobservable.
Note that the relative difference term is not sufficient to alleviate drift in this case. 
This can be seen from the full probabilistic model without marginalization in which the parameters could drift consistently across all frames in the unobservable dimensions.
The corresponding  objective function is summarized as $E_{\mathrm{param}} = \mathbf{r}_{\mathbf{p}_{\mathrm{dyn,rel}}}^\top \mathbf{\Sigma}_{\mathbf{p}_{\mathrm{dyn,rel}}}^{-1} \mathbf{r}_{\mathbf{p}_{\mathrm{dyn,rel}}} + 
\mathbf{r}_{\mathbf{p}_{\mathrm{dyn,abs}}}^\top \mathbf{\Sigma}_{\mathbf{p}_{\mathrm{dyn,abs}}}^{-1} \mathbf{r}_{\mathbf{p}_{\mathrm{dyn,abs}}}$, where $\mathbf{\Sigma}_{\mathbf{p}_{\mathrm{dyn,rel}}}^{-1}$ and 
$\mathbf{\Sigma}_{\mathbf{p}_{\mathrm{dyn,abs}}}^{-1}$ are the diagonal weight matrices.
In summary, the overall objective function of our dynamics augmented VIO is
$
    E_{\mathrm{st-vio}} = E_{\mathrm{vio}} + E_{\mathrm{marg}} + E_{\mathrm{dyn}} + E_{\mathrm{geom}} + E_{\mathrm{extr}} + E_{\mathrm{param}},
$
where $E_{\mathrm{vio}}$ and $E_{\mathrm{marg}}$ are the terms for the visual-inertial odometry and the marginalization prior (see~\cite{usenko19nfr}).
The VIO system for planar motion is ill-posed and the accelerometer bias is unobservable if there is no rotation~\cite{wu_vinsonwheels}. Therefore, the dynamics factor should only be used when the accelerometer bias is converged. We approximate the variance of accelerometer bias $\mathbf{b}_a$ by inverting the related Hessian matrix part.
The dynamics factor is integrated when the variance of the accelerometer bias $\mathbf{b}_{a,t_0}$ at the first active recent frame is smaller than a threshold.

\subsection{Offline Initial Guess Estimation}
\label{sec:init}
The dynamics augmented VIO requires a reasonable initialization of the parameters and extrinsics to guarantee that the numerical solver of the ODE can output a plausible solution. 
We determine the center of mass position of our mobile robot and initialize the extrinsics~$^{\mathrm{o}}\mathbf{T}_{\mathrm{i}}$ from the CAD model.
The initial guess of steering ratio $\gamma$ is approximated by the ratio between the max. value of steering control and the max. front wheel steering angle. 
%
The remaining parameters like the throttle mapping and tire coefficient are agnostic and thus an initial guess is found through offline optimization. 
We first manually select reasonable hyper-parameters $\{ \psi, \tau, \sigma \}$ in the longitudinal force function and only optimize for $C_{\mathrm{thr},1/2}$, $C_{\mathrm{res}}$ with pure forward motion data.
Once the longitudinal force related parameters are identified, we optimize for all hyper-parameters, dynamics model parameters and extrinsics together using data collected with various steering inputs mixed with stop-and-go motion based on the dynamics factor and geometry constraints introduced in the previous section.
During online optimization the hyper-parameters $\{ \psi, \tau, \sigma \}$ are fixed. 
Since the extrinsics is time-varying, we still use the CAD model estimate as initial guess of the neutral position of the suspension for the online optimization phase.

\section{EXPERIMENTS}
\begin{figure}[tb!]
	\centering
	\subfloat{\includegraphics[width=.6\linewidth]{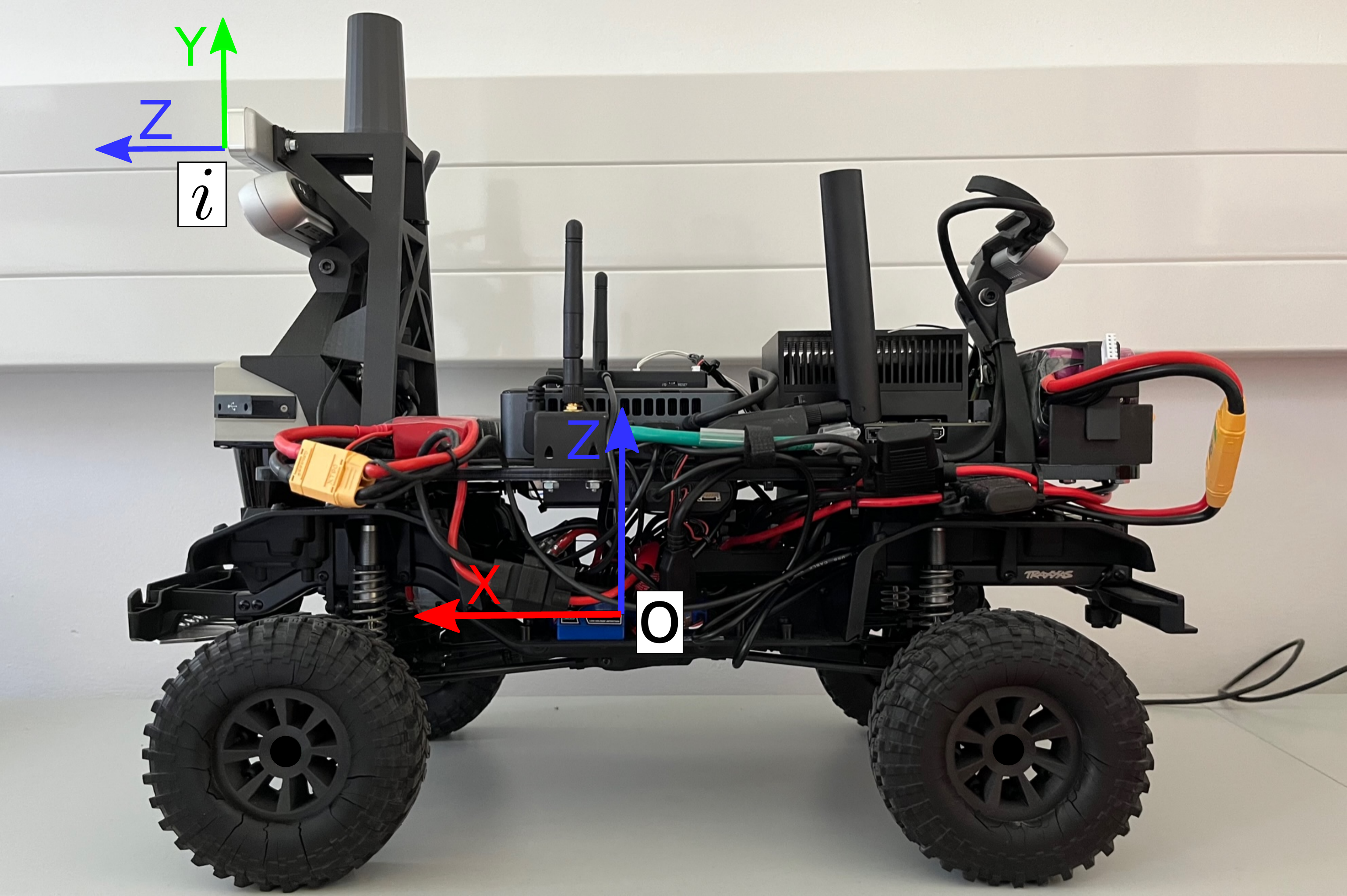}} 
	\subfloat{\includegraphics[width=.303\linewidth]{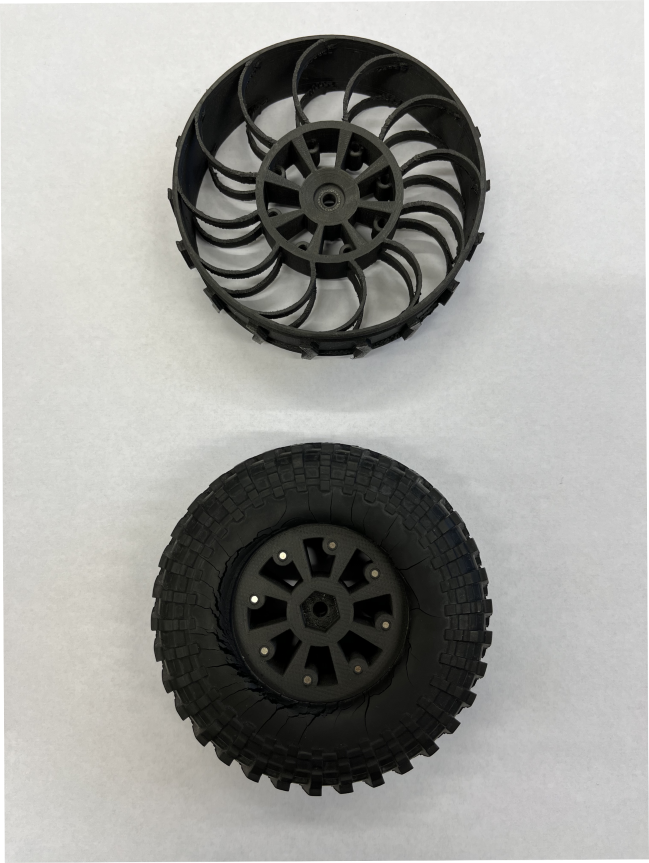}} 
    \caption{Left: Our mobile robot is a modified 1/10 electric RC car equipped with an Intel Realsense T265 stereo camera.
    Right: We primarily use the bottom wheel in our experiments and also evaluate with wheels without rubber tire (top).}
	\label{fig:vehicle}
\end{figure}

We evaluate our proposed method with real-world data collected by our robot (Fig.~\ref{fig:vehicle}). It is a 1/10 scale electric car equipped with a Realsense T265 stereo-fisheye camera with built-in IMU. We control the robot manually and record the images at 30\,Hz, IMU measurements at 200\,Hz and control inputs at 20\,Hz.
We evaluate tracking and prediction accuracy in various scenarios and compare with the original VIO.
We use the same settings for pure VIO and our method with 3 active recent and 7 keyframes in the window.
Similar like~\cite{lee_onlincalib,wu_vinsonwheels},
we use the results of global mapping as ground truth, where we set every frame as keyframe and perform dense bundle adjustment for accuracy and consistency. 
For all experiments, the weight values are set to $10^{3}$, $10^4$, $1$, $30\,\mathrm{d}t \times 10^{4}$, $20$, and $30\,\mathrm{d}t \times 10^{8}$ for the dynamics, geometry constraint, extrinsics initial prior, extrinsics random walk, dynamics parameters absolute prior, and dynamics parameters random walk factor, respectively, where $\mathrm{d}t$ is the time interval between two frames. Parameters $\psi, \tau$ and $\sigma$ are found as 0.202, 2.335 and 10 by offline optimization.
The variance threshold for the accelerometer bias is $4.5\times 10^{-4}$.

\subsection{Experiment Setup}
To evaluate tracking and prediction accuracy and the calibration capability, we collect data in both indoor and outdoor scenes. Each scene contains two different groups of terrain and each group has three data sequences. The indoor scenes include a lobby with tile floor and a corridor with concrete floor. In contrast to the indoor scene,
the outdoor scene is not perfectly flat with small bumpiness and contains places with concrete surface and paved brick road. For each scene, we control our robot with various steering inputs and either full or varying throttle inputs. 
In all data recordings, the robot starts from a static pose.
The mass $m$ and wheel distance $l_f + l_r$ for our robot are measured directly. The  momentum of inertia $I_{\mathrm{z}}$ and the distance between center of mass and front wheel $l_f$ are approximated from CAD software.
We then perform the offline initialization strategy described in section \ref{sec:init} to initialize the dynamics model parameters~$\mathbf{p}_{\mathrm{dyn}}$ using two short trajectories of 10\,s in the corridor scene. 

\subsection{Tracking Accuracy Evaluation}
We evaluate the tracking accuracy of our ST-VIO by comparing the relative pose error~(RPE)~\cite{Zhang18iros} with the original VIO, to show the relative improvement. Note that since no previous method is available that optimizes a single-track dynamics model with VIO, we can only compare our approach with the baseline VIO in our experiments. 
The comparison between our base VIO and other popular VIO methods can be found in~\cite{usenko19nfr, orbslam3}.
The RPE value is generated by computing the errors over $10,20,...,50\%$ sequence lengths of the full trajectory. We exclude the standing still segment at the end of the trajectories to avoid biasing the tracking error to low values in this trivial case.
Table~\ref{tab:traj_rpe} provides average results over all indoor and outdoor sequences.  
For the indoor data, our approach~{ST-VIO} overall improves trajectory accuracy. 

The outdoor data are challenging for our method, as the bumpy terrain could violate the single-track dynamics model. In most outdoor sequences, our method can still improve the accuracy. In the concrete data group with varying throttle, the rotational accuracy drops slightly. Besides the less even terrain, another reason could be that the vehicle speed in the varying throttle case is relatively small comparing to the full throttle case and integrating the dynamics model cannot improve the accuracy further.
%
%
\begin{table}[tb!]
	\caption{Average trajectory RPE on indoor and outdoor sequences ({ST-VIO}: ours, {VIO}: pure VIO, \emph{-full}: full throttle maneuver, \emph{-varying}: varying throttle maneuver).
	}
	\label{tab:traj_rpe}
	\begin{center}
		\begin{tabular}{ccccc}
			\toprule
			& \multicolumn{2}{c}{transl. RMSE RPE\,[m]} &
			\multicolumn{2}{c}{rot. RMSE RPE\,[deg]}\\
			\cmidrule(lr){2-3} \cmidrule(lr){4-5}
			dataset  & {VIO} & {ST-VIO} & {VIO} & {ST-VIO}\\
			\midrule
            \textit{lobby-full} & 0.118 & \textbf{0.108} & 1.633 & \textbf{1.573} \\
            \textit{lobby-varying} & 0.076 & \textbf{0.069} & 1.038 & \textbf{1.006} \\
            \textit{corridor-full} & 0.183 & \textbf{0.174} & 0.993 & \textbf{0.941} \\
            \textit{corridor-varying} & 0.120 & \textbf{0.114} & 0.675 & \textbf{0.659} \\
			\midrule
            \textit{concrete-full} & 0.176 & \textbf{0.162} & 1.500 & \textbf{1.423} \\
            \textit{concrete-varying} & 0.168 & \textbf{0.164} & \textbf{1.192} & 1.195 \\
            \textit{brick-full} & 0.108 & \textbf{0.107} & 1.220 & \textbf{1.200} \\
            \textit{brick-varying} & 0.116 & \textbf{0.108} & 0.764 & \textbf{0.748} \\
			\bottomrule
		\end{tabular}
	\end{center}
\end{table}
We also perform an ablation study for tracking accuracy evaluation where only geometry constraints are applied, and the dynamics factor is deactivated. 
VIO with only geometry constraints shows similar accuracy of 0.133\,m and 1.126\,$\deg$ like the original VIO while our approach achieves 0.124\,m and 1.093\,$\deg$ RSME of transl. and rot. RPE in average for all data sequences. 
The algorithm with dynamics factor diverges on some sequences without geometry constraints.

\subsection{Prediction Accuracy Evaluation}
We also evaluate prediction accuracy for different time horizons (0.33\,s, 0.66\,s, 1.66\,s, 3.33\,s and 10\,s) to validate the online calibration of the parameters. The prediction is computed with the current dynamics model parameters from the start state estimated by ST-VIO at each frame. 
The prediction is a relative 2D pose in the vehicle body frame. 
To compare with ground truth camera poses, we project the relative camera pose to the vehicle body frame using the optimized extrinsics. 
The standing still part at the end of the trajectories are excluded again to avoid including the perfect but trivial predictions~(zero relative pose and velocity) into the evaluation. 
RPE results are summarized in Table~\ref{tab:pred_rpe}.
The prediction accuracy is denoted as \emph{calib} and \emph{init} using online calibrated and initial parameters, respectively.
We observed improved prediction accuracy using the online calibrated parameters across all sequences. For the corridor sequences, the improvement is relatively modest. This is attributed to the fact that offline optimization is performed on sequences captured within the corridor scene.
We refer to the supplementary video for a visualization of the predictions.

\begin{table}[tb!]
	\caption{Average prediction RPE on indoor and outdoor sequences (\emph{init}: offline-calibrated, \emph{calib}: online-calibrated, \mbox{\emph{-full}}: full throttle, \emph{-varying}: varying throttle maneuver).
	}
	\label{tab:pred_rpe}
	\begin{center}
		\begin{tabular}{ccccc}
			\toprule
			& \multicolumn{2}{c}{transl. RMSE RPE\,[m]} &
			\multicolumn{2}{c}{rot. RMSE RPE\,[deg]}\\
			\cmidrule(lr){2-3} \cmidrule(lr){4-5}
			dataset  & \emph{init} & \emph{calib} & \emph{init} & \emph{calib}\\
			\midrule
            \textit{lobby-full} & 1.120 & \textbf{0.453} & 22.383 & \textbf{7.558} \\
            \textit{lobby-varying} & 0.764 & \textbf{0.477} & 9.588 & \textbf{5.790} \\
            \textit{corridor-full} & 0.550 & \textbf{0.524} & 7.367 & \textbf{6.128} \\
            \textit{corridor-varying} & 0.647 & \textbf{0.552} & 6.092 & \textbf{4.883} \\
			\midrule
            \textit{concrete-full} & 1.962 & \textbf{0.508} & 25.884 & \textbf{6.283} \\
            \textit{concrete-varying} & 1.007 & \textbf{0.518} & 12.191 & \textbf{4.202} \\
            \textit{brick-full} & 0.701 & \textbf{0.310} & 12.077 & \textbf{3.781} \\
            \textit{brick-varying} & 0.625 & \textbf{0.496} & 8.613 & \textbf{5.493} \\
			\bottomrule
		\end{tabular}
	\end{center}
\end{table}

\begin{figure}[tb!]
	\centering
	\subfloat{\includegraphics[width=.57\linewidth]{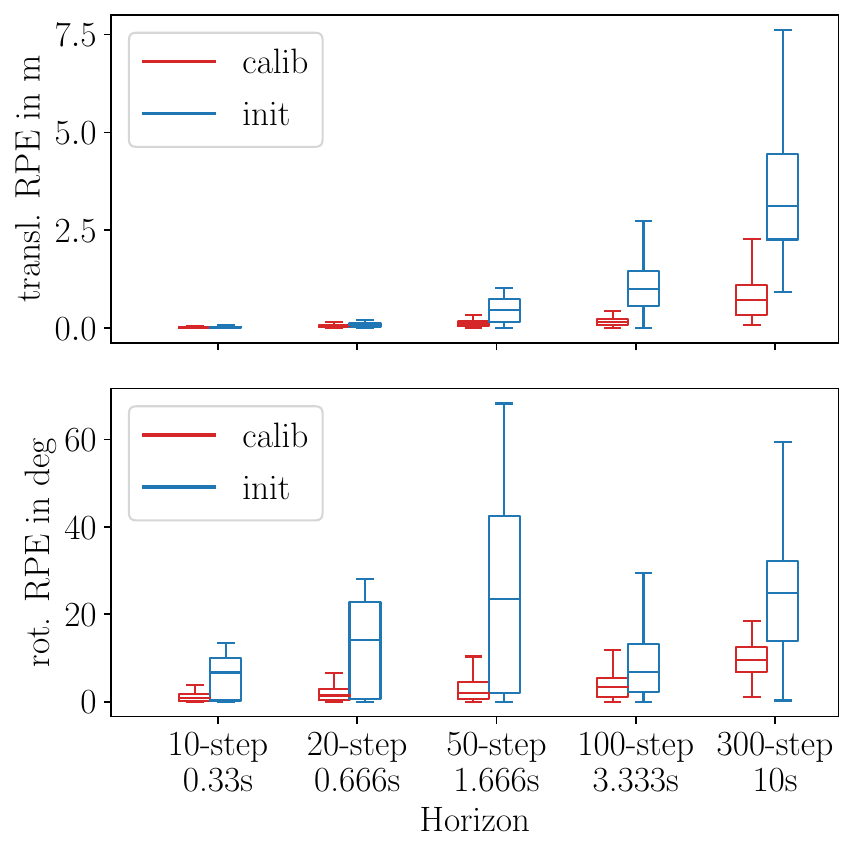}} \hspace{1ex}
	\subfloat{\includegraphics[width=.39\linewidth]{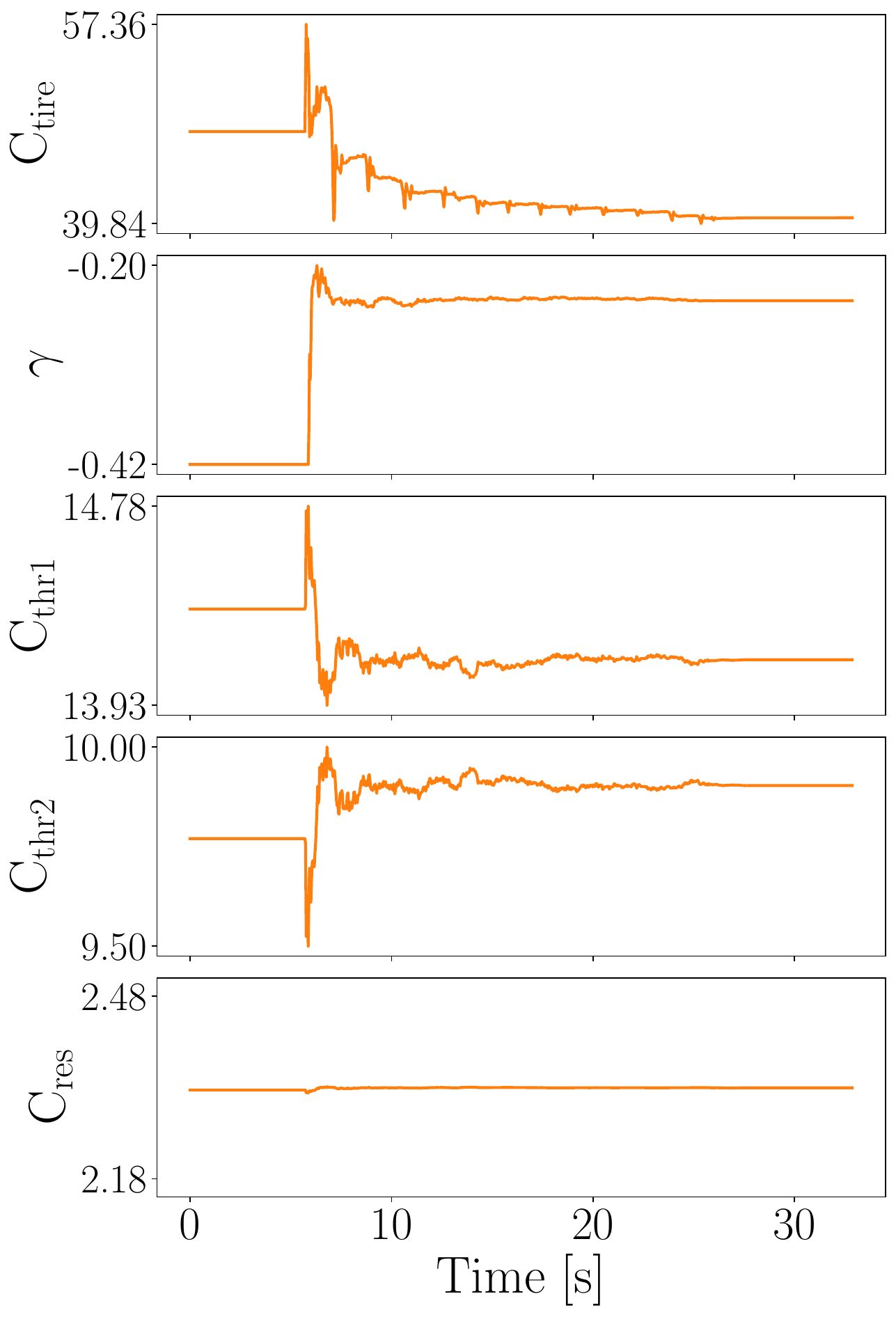}} \\
	\subfloat{\includegraphics[width=.57\linewidth]{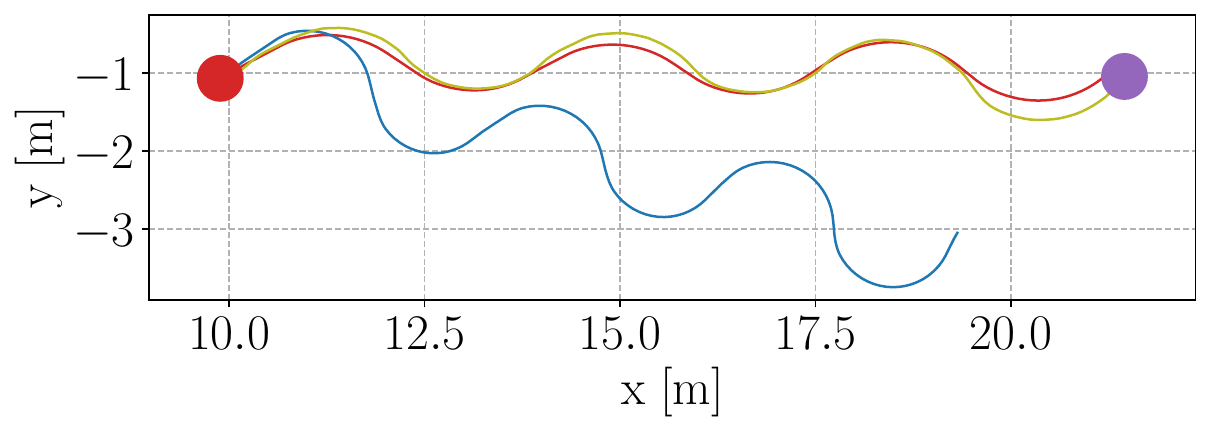}}
	\subfloat{\includegraphics[width=.41\linewidth]{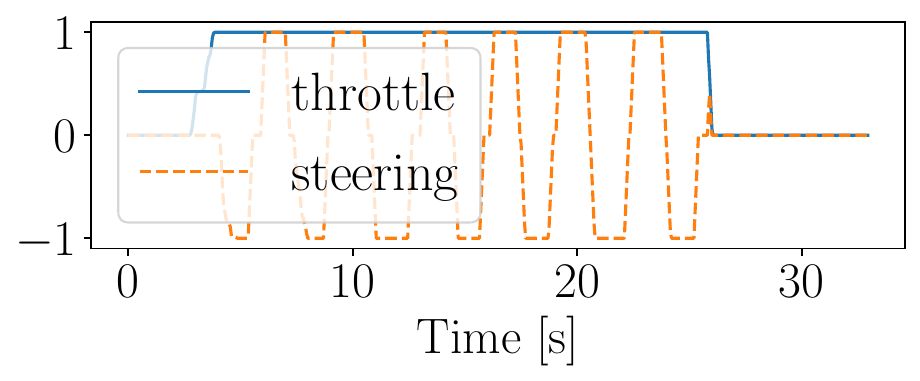}}
    \caption{Top left: online calibration (calib) by ST-VIO for the new wheels clearly improves prediction over offline calibration (init) for the old wheels. 
    Top right: evolution of online calibrated parameters (calib).
    Bottom left: 10\,s prediction results (red: calib, blue: init, yellow trajectory: ground truth, red/purple circle: start/end, rotated by $30\degree$ for visualization). 
    Bottom right: control inputs.}
    \label{fig:wheel_pred}
\end{figure}

\paragraph{Change of Robot Properties}
We additionally collect a data sequence with different wheels~(top one in Fig.~\ref{fig:vehicle}) that have lower traction in the corridor scene. 
We run our method using the initial dynamics model parameters for the old set of wheels as in the experiments above.
The bottom image in Fig.~\ref{fig:wheel_pred} illustrates the prediction error qualitatively. 
The top left image in Fig.~\ref{fig:wheel_pred} demonstrate that the online calibrated parameters show significantly less error than the parameters calibrated offline for the old wheel for various time horizons.
The top right figure in Fig.~\ref{fig:wheel_pred} depicts the evolution of the online calibrated parameters during the optimization.
The biggest adaptation is in the tire coefficient $C_{\mathrm{tire}}$ and steering ratio $\gamma$ parameters, while the throttle mapping parameters are only adapted in a relatively small range. 
Please refer to the supplementary video for a visualization of the results.

\paragraph{Stop-and-Go Motion}
We also collect three sequences in each indoor environment for repeated stop-and-go motion with varying steering to demonstrate that our singularity-free formulation  enables calibration and prediction in this case. 
In the lobby scene, the average transl. RMSE RPE improves from 0.405\,m to 0.340\,m,
the average rot. RMSE RPE improves from 11.294\,$\deg$ to 7.874\,$\deg$.
For the corridor sequences, the average transl. RMSE RPE changes from 0.332\,m to 0.387\,m,
the average rot. RMSE RPE improves slightly from 8.812\,$\deg$ to 8.645\,$\deg$. For the corridor sequences, the online calibration does not further enhance prediction accuracy since offline calibration was already conducted on similar sequences. In the lobby scene, our method demonstrates adaptability and improves prediction even under stop-and-go movements. 



\subsection{Run-Time Evaluation}
We evaluate the run-time time of our method compared to the pure VIO on an Intel i9-10900X CPU@3.70GHz with 20 threads. For pure VIO processing, one frame takes 6.15\,ms in average, while our dynamics augmented VIO needs 16.60\,ms. 
Our method needs about three times more run-time than the original VIO yet is still real-time capable since avg. run-time is below the frame interval (33.3\,ms).

\section{Conclusion} 
\label{sec:conclusion}
In this work, we propose ST-VIO
for wheeled robots which integrates a singularity-free single-track vehicle dynamics model and optimizes for the vehicle parameters online together with the VIO states in a sliding window fashion. 
The vehicle model is tightly integrated by introducing a dynamics factor which minimizes the difference between the pose and velocity prediction based on the model and the state estimate. 
A multistep objective function is constructed by predicting the pose and velocity from the first frame until the end frame in the window. 
In experiments, we demonstrate that our method is real-time capable and can
improve the tracking accuracy on flat ground, especially for the motions with full throttle. 
We also demonstrate that online calibration can improve motion prediction and adapt the parameters to changes of the environment and wheel properties. 
In future work, we aim to integrate vehicle models which can handle more complex terrain properties.








\bibliographystyle{IEEEtran}
\bibliography{IEEEabrv, references}

\end{document}